# TrackingNet: A Large-Scale Dataset and Benchmark for Object Tracking in the Wild

Matthias Müller*, Adel Bibi*, Silvio Giancola*,
Salman Al-Subaihi, Bernard Ghanem

King Abdullah University of Science and Technology

**Abstract.** Despite the numerous developments in object tracking, further development of current tracking algorithms is limited by small and mostly saturated datasets. As a matter of fact, data-hungry trackers based on deep-learning currently rely on object detection datasets due to the scarcity of dedicated large-scale tracking datasets. In this work, we present TrackingNet, the first large-scale dataset and benchmark for object tracking in the wild. We provide more than 30K videos with more than 14 million dense bounding box annotations. Our dataset covers a wide selection of object classes in broad and diverse context. By releasing such a large-scale dataset, we expect deep trackers to further improve and generalize. In addition, we introduce a new benchmark composed of 500 novel videos, modeled with a distribution similar to our training dataset. By sequestering the annotation of the test set and providing an online evaluation server, we provide a fair benchmark for future development of object trackers. Deep trackers fine-tuned on a fraction of our dataset improve their performance by up to 1.6% on OTB100 and up to 1.7% on TrackingNet Test. We provide an extensive benchmark on TrackingNet by evaluating more than 20 trackers. Our results suggest that object tracking in the wild is far from being solved.

## 1 Introduction

Object tracking is a common task in computer vision, with a long history spanning decades [1–3]. Despite considerable progress in the field, object tracking remains a challenging task. Current trackers perform well on established datasets such as OTB [4, 5] and VOT [6–11] benchmarks. However, most of these datasets are fairly small and do not fully represent the challenges faced when tracking objects *in the wild*.

Following the rise of deep learning in computer vision, the tracking community is currently embracing data-driven learning methods. Most trackers submitted to the annual challenge VOT17 [11] use deep features, while they were nonexistent in earlier versions VOT13 [7] and VOT14 [8]. In addition, nine out of the ten top-performing trackers in VOT17 [11] rely on deep features, outperforming the previous state-of-the-art trackers. However, the tracking community still lacks a dedicated large-scale dataset to train deep trackers. As a consequence, deep trackers are often restricted to using pretrained models from object classification [12] or use object detection datasets such as ImageNet Videos [13]. As an example of this, SiameseFC [14] and CFNet [15] show outstanding results by training specific Convolutional Neural Networks (CNNs) for tracking.

* equal contribution

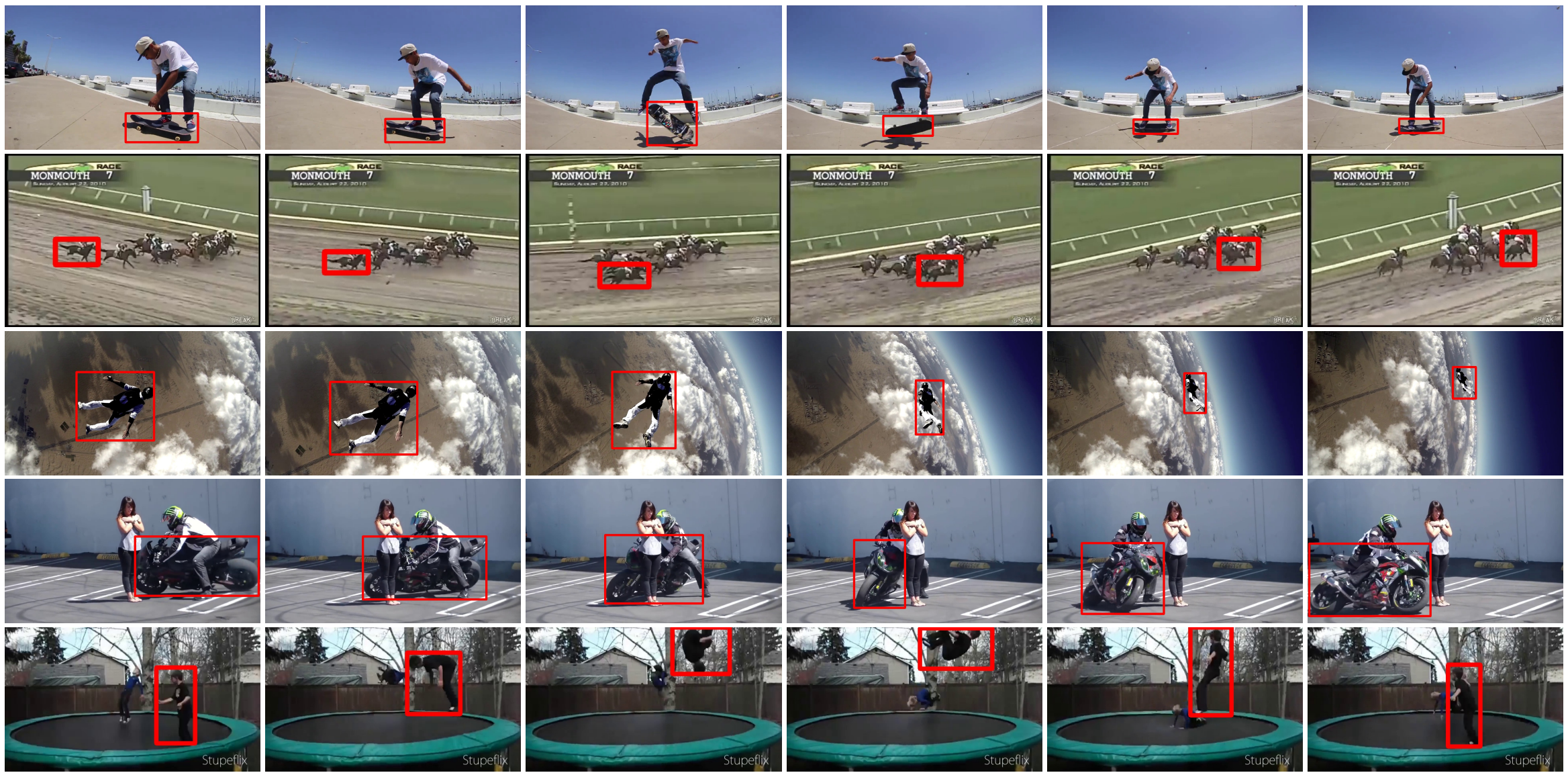

Fig. 1: Examples of tracking from our novel TrackingNet Test set.

Since classical trackers rely on handcrafted features and because existing tracking datasets are small, there is currently no clear split between data used for training and testing. Recent benchmarks [11, 16] now consider putting aside a sequestered test set to provide a fair comparison. Yet, these test sets are small and not dedicated for training purposes. Hence, it is common to see trackers developed and trained on the OTB [5] dataset before competing on VOT [6]. Note that VOT15 [9] is sampled from existing datasets like OTB100 [5] and ALOV300 [17], resulting in overlapping sequences (*e.g.* basketball, car, singer, *etc...*). Even though the redundancy is contained, one needs to be careful while selecting training video sequences, since training deep trackers on testing videos is not fair. As a result, there is usually not enough data to train deep networks for tracking and data from different fields are used to pre-train models, which is a limiting factor for certain architectures.

In this paper, we present TrackingNet, a large-scale object tracking dataset designed to train deep trackers. Our dataset has several advantages. First, the large training set enables the development of deep design specific for tracking. Second, the specificity of the dataset for object tracking enables novel architectures to focus on the temporal context between consecutive frames. Current large scale object detection datasets do not provide data densely annotated in time. Third, TrackingNet represents real-world scenarios by sampling over YouTube videos. As such, TrackingNet videos contain a rich distribution of object classes, which we enforce to be shared between training and testing. Last, we evaluate tracker performance on a segregated testing set with a similar distribution over object classes and motion. Trackers do not have access to the annotations of these videos but can obtain results and insights through an evaluation server.

**Contributions. (i)** We present TrackingNet, the first large-scale dataset for object tracking. We analyze the characteristics, attributes and uniqueness of TrackingNet when compared with other datasets (Section 3). **(ii)** We provide insights into different techniques to generate dense annotations from coarse ones. We show that most trackers can produce accurate and reliable dense annotations over 1 second-long intervals. (Section 4). **(iii)** We provide an extended baseline for state-of-the-art trackers benchmarked on TrackingNet. We show that pre-training deep models on TrackingNet can improve their performance on other datasets by increasing their metrics by up to 1.7%. (Section 5).

## 2 Related Work

In the following, we provide an overview of the various research on object tracking. The tasks in the field can be clustered between *multi-object tracking* [5, 6] and *single-object tracking* [18, 16]. The former focuses on multiple instance tracking of class-specific objects, relying on strong and fast object detection algorithms and association estimation between consecutive frames. The latter is the target of this work. It approaches the problem by *tracking-by-detection*, which consists of two main components: *model representation*, either generative [19, 20] or discriminative [21, 22], and *object search*, a trade-off between computational cost and dense sampling of the region of interest.

**Correlation Filter Trackers.** In recent years, correlation filter (CF) trackers [23–26] have emerged as the most common, fastest and most accurate category of trackers. CF trackers learn a filter at the first frame, which represents the object of interest. This filter localizes the target in successive frames before being updated. The main reason behind the impressive performance of CF trackers lies in the approximate dense sampling achieved by circulantly shifting the target patch samples [24]. Also, the remarkable runtime performance is achieved by efficiently solving the underlying ridge regression problem in the Fourier domain [23].

Since the inception of CF trackers with single-channel features [23, 24], they have been extended with kernels [25], multi-channel features [27] and scale adaptation [28]. In addition, many works enhance the original formulation by adapting the regression target [29], adding context [30, 31], spatially regularizing the learned filters and learning continuous filters [32].

**Deep Trackers.** Beside the CF trackers that use deep features from object detection networks, few works explore more complete deep learning approaches. A first approach consists of learning generic features on a large-scale object detection dataset and successively fine-tuning domain-specific layers to be target-specific in an online fashion. MDNET [33] shows the success of such a method by winning the VOT15 [9] challenge. A second approach consists of training a fully convolutional network and using a feature map selection method to choose between shallow and deep layers during tracking [34]. The goal is to find a good trade-off between general semantic and more specific discriminative features, as well as, to remove noisy and irrelevant feature maps.

While both of these approaches achieve state-of-the-art results, their computation cost prohibits these algorithms from being deployed in real applications. A third approach consists of using Siamese networks that predict motion between consecutive frames. Such trackers are usually trained offline on a large-scale dataset using either deep regression [35] or a CNN matching function [14, 15, 36]. Due to their simple architecture and lack of online fine-tuning, only a forward pass has to be executed at test time. This results in very fast run-times (up to 100fps on a GPU) while achieving competitive accuracy. However, since the model is not updated at test time, the accuracy highly depends on how well the training dataset captures appearance nuisances that occur while tracking various objects. Such approaches would benefit from a large-scale dataset like the one we propose in this paper.

**Object Tracking Datasets.** Numerous datasets are available for object tracking, the most common ones being OTB [5], VOT [6], ALOV300 [17] and TC128 [37] for single-object tracking and MOT [18, 16] for multi-object tracking. **VIVID** [38] is an early attempt to build a tracking dataset for surveillance purposes. **OTB50** [4] and **OTB100** [5] provide 51 and 98 video sequences annotated with 11 different attributes and upright bounding boxes for each frame. **TC128** [37] comprises 129 videos, based on similar attributes and upright bounding boxes. **ALOV300** [17] comprises 314 videos sequences labelled with 14 attributes. **VOT** [6] proposes several challenges with up to 60 video sequences. It introduced rotated bounding boxes as well as extensive studies on object tracking annotations. **VOT-TIR** is a specific dataset from VOT focusing on Thermal InfraRed videos. **NUS PRO** [39] gathers an application-specific collection of 365 videos for people and rigid object tracking. **UAV123** and **UAV20L** [40] gather another application-specific collection of 123 videos and 20 long videos captured from a UAV or generated from a flight simulator. **NfS** [41] provides a set of 100 videos with high framerate, in an attempt to focus on fast motion. Table 1 provides a detailed overview of the most popular tracking datasets.

Despite the availability of several datasets for object tracking, large scale datasets are necessary to train deep trackers. Therefore, current deep trackers rely on object detection datasets such as ImageNet Video [13] or Youtube-BoundingBoxes [42]. Those datasets provide object detection bounding boxes on videos, relatively sparse in time or at a low frame rate. Thus, they lack motion information about the object dynamics in consecutive frames. Still, they are widely used to pre-train deep trackers. They provide deep feature representation with object knowledge that can be transferred from detection to tracking.

## 3 TrackingNet

In this section, we introduce TrackingNet, a large-scale dataset for object tracking. TrackingNet assembles a total of 30,643 video segments with an average duration of 16.6s. All the 14,431,266 frames extracted from the 140 hours of visual content are annotated with a single upright bounding box. We provide a comparison with other tracking datasets in Table 1 and Figure 2.

Table 1: Comparison of current datasets for object tracking.

| Datasets | Nb Videos | Nb Annot. | Frame per Video | Nb Classes |
|---|---|---|---|---|
| **VIVID** [38] | 9 | 16274 | 1808.2 | - |
| **TC128** [37] | 129 | 55652 | 431.4 | - |
| **OTB50** [4] | 51 | 29491 | 578.3 | - |
| **OTB100** [5] | 98 | 58610 | 598.1 | - |
| **VOT16** [10] | 60 | 21455 | 357.6 | - |
| **VOT17** [11] | 60 | 21356 | 355.9 | - |
| **UAV20L** [40] | 20 | 58670 | 2933.5 | - |
| **UAV123** [40] | 91 | 113476 | 1247.0 | - |
| **NUS PRO** [39] | 365 | 135305 | 370.7 | - |
| **ALOV300** [17] | 314 | 151657 | 483.0 | - |
| **NfS** [36] | 100 | 383000 | 3830.0 | - |
| **MOT16** [16] | 7 | 182326 | 845.6 | - |
| **MOT17** [16] | 21 | 564228 | 845.6 | - |
| **TrackingNet (Train)** | **30132** | **14205677** | **471.4** | **27** |
| **TrackingNet (Test)** | **511** | **225589** | **441.5** | **27** |

Our work attempts to bridge the gap between data-hungry deep trackers and scarcely-available large scale datasets. Our proposed tracking dataset is larger than the previous largest one by 2 orders of magnitude. We build TrackingNet to address object tracking in the wild. Therefore, the dataset copes with a large variety of frame rates, resolutions, context and object classes. In contrast with previous tracking datasets, TrackingNet is split between training and testing. We carefully select 30,132 training videos from Youtube-BoundingBoxes [42] and build a novel set of 511 testing videos with a distribution similar to the training set.

### 3.1 From YT-BB to TrackingNet Training Set

Youtube-BoundingBoxes (YT-BB) [42] is a large scale dataset for object detection. This dataset consists of approximately 380,000 video segments, annotated every second with upright bounding boxes. Those videos are gathered directly from YouTube, with a wide diversity in resolution, frame rate and duration.

Since YT-BB focuses on object detection, the object class is provided along with the bounding boxes. The dataset proposes a list of 23 object classes representative of the videos available on the YouTube platform. For the sake of tracking, we remove the object classes that lack motion by definition, in particular *potted plant* and *toilet*. Since the *person* class represents 25% of the annotations, we split it into 7 different classes based on their context. Overall, the distribution of the object classes in TrackingNet is shown in Figure 3.

To ensure decent quality in the videos for tracking purposes, we filtered out 90% of the the videos based on attribute criteria. First, we avoid small segments by removing videos shorter than 15 seconds. Second, we only considered bounding boxes that covered less than 50% of the frame. Last, we preserve segments that contain at least a reasonable amount of motion between bounding boxes. During such filtering, we preserved the original distribution of the 21 object classes provided by YT-BB, to prevent bias in the dataset. We end up with a training set of 30,132 videos, which we split into 12 training subsets, each of which contains 2,511 videos and preserves the original YT-BB object classes distribution.

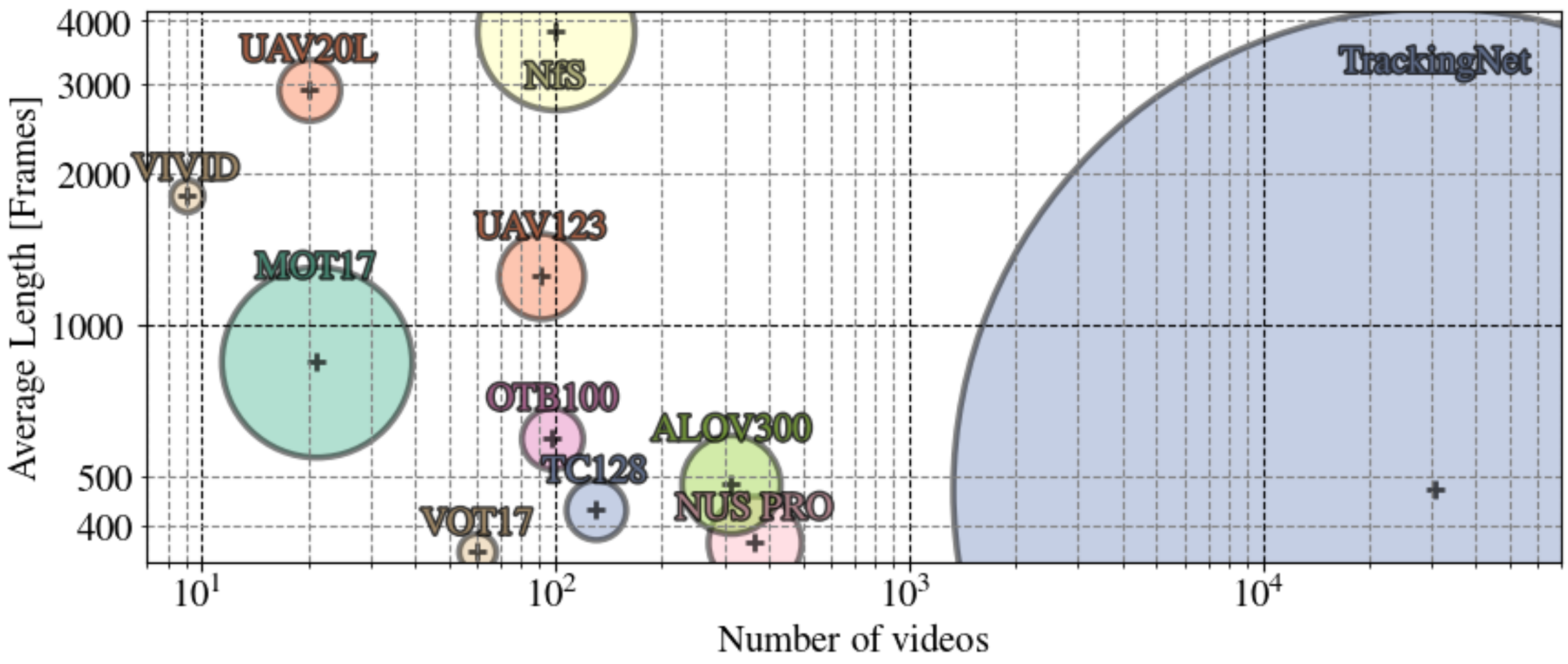

Fig. 2: Comparison of tracking datasets distributed across the number of videos and the average length of the videos. The size of circles is proportional to the number of annotated bounding boxes. Our dataset has the largest amount of videos and frames and the video length is still reasonable for short video tracking.

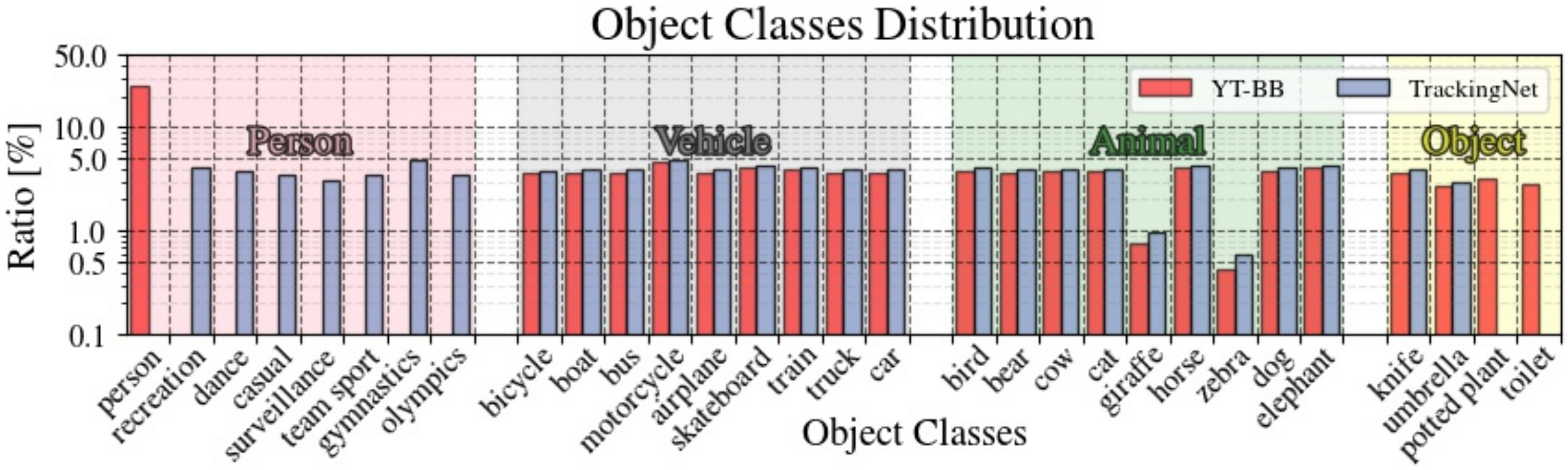

Fig. 3: Definition of object classes and macro classes.

Coarse annotations are provided by YT-BB at 1 fps. In order to increase the annotation density, we rely on a mixture of state-of-the-art trackers to fill in missing annotations. We claim that any tracker is reliable on a small time lapse of 1 second. We present in Section 4 the performance of state-of-the-art trackers on 1 second-long video segments from OTB100. As a result, we densely annotated the 30,132 videos using a weighted average between a forward and a backward pass using the DCF tracker [25].

By doing so, we provide a densely annotated training dataset for object tracking, along with code for automatically downloading videos from YouTube and extracting the annotated frames.

### 3.2 From YT-CC to TrackingNet Testing Set

Alongside the training dataset, we compile a novel dataset for testing, which comprises 511 videos from YouTube with Creative Commons licence, namely YT-CC. We carefully select those videos to reflect the object class distribution from the training set. We ensure that those videos do not contain any copyrights, so they can be shared.

We used Amazon Mechanical Turk workers (Turkers) for annotating those those videos. We annotate the first bounding boxes and define specific rules for the Turkers to carefully annotate the successive frames. We define the objects as in YT-BB for object detection, *i.e.* with the smallest bounding box fitting any visible part of the object to track.

Annotations should be defined in a deterministic way, using rules that are agreed upon and abided by during the annotation process. By defining the smallest upright bounding box around an object, we avoid any ambiguity. However, the bounding box may contain a large amount of background. For instance, the arm and the legs are always included for the *person* class, regardless of the person's pose. We argue that a tracker should be able to cope with deformable objects and to understand what it is tracking. In a similar fashion, the tails of animal are always included. In addition, the bounding box of an object is adjusted as a function of its visibility in the frame. Estimating the position of an occluded part of the object is not deterministic hence should be avoided. For instance, the handle of the object class *knife* could be hidden by the hand. In such cases, only the blade is annotated.

We use the VATIC tool [43] to annotate the frames. It incorporates an optical flow algorithm to guess the position of the next bounding boxes in successive frames. Turkers may annotate a non-tight bounding box around the object or rely on the optical flow to determine the bounding box location and size. To avoid such behavior, we visually inspect every single frame after each annotation round, rewarding good Turkers and rejecting bad annotations. We either restart the video annotation from scratch or ask Turkers to fine-tune previous results. With our supervision in the loop, we ensure the quality of our annotations after a few iterations, discourage bad annotators and incentivize the good ones.

### 3.3 Attributes

Successively, each video is annotated with a list of attributes defined in Table 2. 15 attributes are provided for our testing set, the first 5 are extracted automatically by analyzing the variation of the bounding boxes in time while the last 10 are manually checked by visually analyzing the 511 videos of our dataset. An overview of the attribute distribution is given in Figure 4 and compared to OTB100 [5] and VOT17 [11].

Table 2: List and description of the 15 attributes that characterize videos in TrackingNet. **Top:** automatically estimated. **Bottom:** visually inspected.

| Attr | Description |
|---|---|
| **SV** | Scale Variation: the ratio of bounding box area is outside the range $[0.5, 2]$ after 1s. |
| **ARC** | Aspect Ratio Change: the ratio of bounding box aspect ratio is outside the range $[0.5, 2]$ after 1s. |
| **FM** | Fast Motion: the motion of the ground truth bounding box is larger than the size of the bounding box. |
| **LR** | Low Resolution: at least one ground truth bounding box has less than 1000 pixels. |
| **OV** | Out-of-View: some portion of the target leaves the camera field of view. |
| **IV** | Illumination Variation: the illumination of the target changes significantly. |
| **CM** | Camera Motion: abrupt motion of the camera. |
| **MB** | Motion Blur: the target region is blurred due to the motion of target or camera. |
| **BC** | Background Clutter: the background near the target has similar appearance as the target. |
| **SOB** | Similar Object: there are objects of similar shape or same type near the target. |
| **DEF** | Deformation: non-rigid object deformation. |
| **IPR** | In-Plane Rotation: the target rotates in the image plane. |
| **OPR** | Out-of-Plane Rotation: the target rotates out of the image plane. |
| **POC** | Partial Occlusion: the target is partially occluded. |
| **FOC** | Full Occlusion: the target is fully occluded. |

First, we claim to have better control over the number of frames per video in our dataset, with a more contained variation with respect to other datasets. We argue that such contained length diversity is more suitable for training with a constant batch size. Second, the distribution of the bounding box resolution is more diverse in TrackingNet, providing more diversity in the scale of the object to track. Third, we show that challenges in OTB100 [5] and VOT17 [11] focus on objects with slightly larger motion, while TrackingNet shows a more natural motion distribution over the fastest moving instances in YT-BB. Similar conclusions can be drawn from the distribution of the aspect ratio change attribute. Fourth, more than 30% of the OTB100 instances have a constant aspect ratio, while VOT17 shows a flatter distribution. Once again, we argue that TrackingNet contains a more natural distribution of objects present in the wild. Last, we show statistics over the 15 attributes, which will be used to generate attribute specific tracking results in Section 5. Overall, we see that our sequestered testing set has an attribute distribution similar to that of our training set.

### 3.4 Evaluation

Annotation for the testing set should not be revealed to ensure a fair comparison between trackers. We thus evaluate the trackers through an online server. In a similar OTB100 fashion, we perform a One Pass Evaluation (OPE) and measure the success and precision of the trackers over the 511 videos. The success $S$ is measured as the Intersection over Union (IoU) of the pixels between the ground truth bounding boxes ($BB^{gt}$) and the ones generated by the trackers ($BB^{tr}$). The trackers are ranked using the Area Under the Curve (AUC) measurement [5]. The precision $P$ is usually measured as the distance in pixels between the centers $C^{gt}$ and $C^{tr}$ of the ground truth and the tracker bounding box, respectively. The trackers are ranked using this metric with a conventional threshold of 20 pixels.

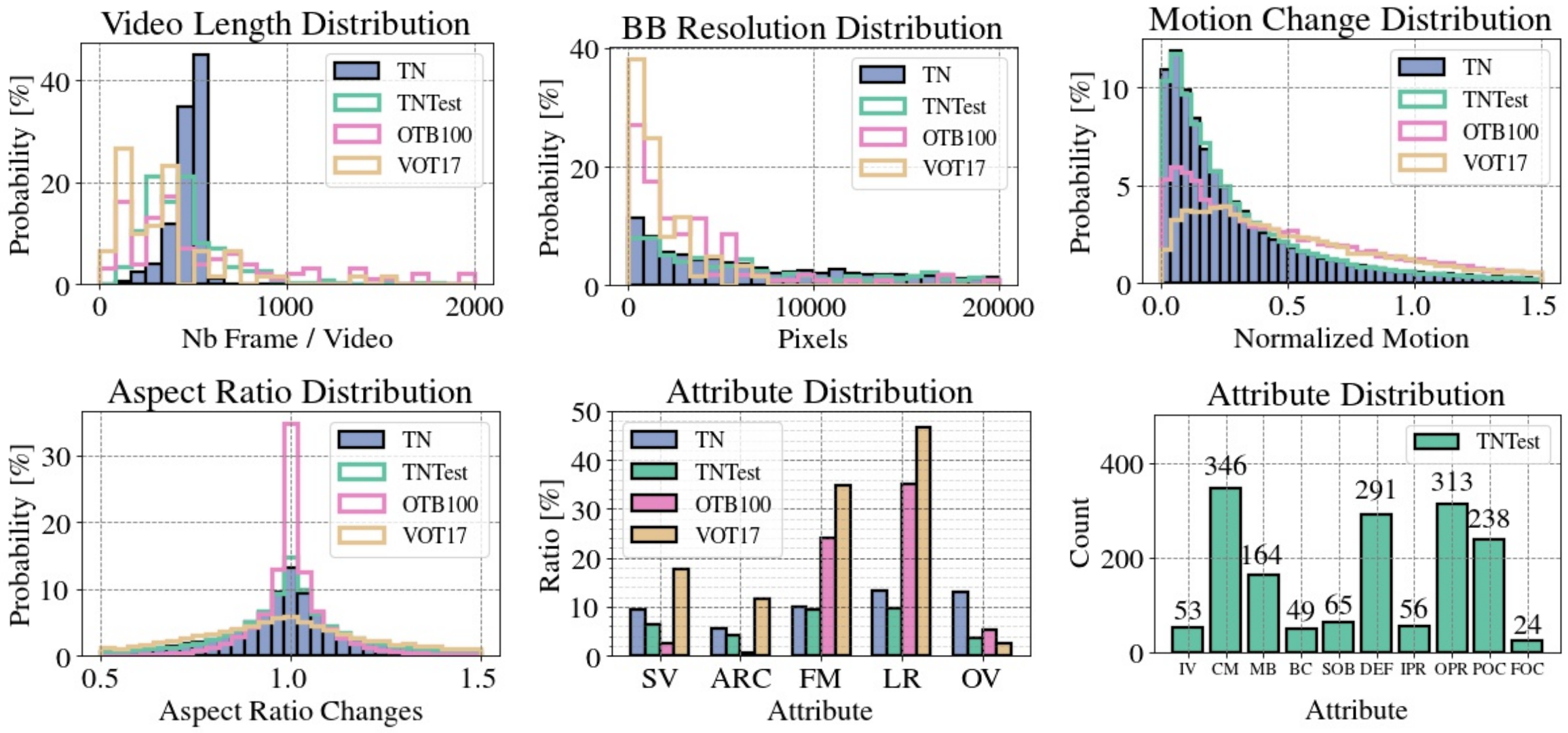

Fig. 4: **(top to bottom, left to right):** Distribution of the tracking videos in term of *Video length*, *BB Resolution*, *Motion Change*, *Scale Variation* and *attributes distribution* for the main tracking datasets.

Since the precision metric is sensitive to the resolution of the images and the size of the bounding boxes, we propose a third metric $P_{norm}$. We normalize the precision over the size of the ground truth bounding box, following Eq. 1. The trackers are then ranked using the AUC for normalized precision between 0 and 0.5. By substituting the original precision with the normalized one, we ensure the consistency of the metrics across different scales of objects to track. However, for bounding boxes with similar scale, success and normalized precision are very similar and show how far an annotation is from another. Nevertheless, we argue that they will differ in the case of different scales. For the sake of consistency, we provide results using precision, normalized precision and success.

$$
\begin{aligned}
S &= \frac{|BB^{tr} \cap BB^{gt}|}{|BB^{tr} \cup BB^{gt}|} \qquad P = \|C^{tr} - C^{gt}\|_2 \\
P_{norm} &= \|W\big(C^{tr} - C^{gt}\big)\|_2 \quad W = \mathrm{diag}(BB^{gt}_x, BB^{gt}_y)
\end{aligned} \tag{1}
$$

## 4 Dataset Experiments

Since TrackingNet Training Set ($\sim$ 30K videos) is compiled from the YT-BT dataset, it is originally annotated with bounding boxes every second. While such sparse annotations might be satisfactory for some vision tasks, *e.g.* object classification and detection, deep network based trackers rely on learning the temporal evolution of bounding boxes over time. For instance, Siamese-like architectures [34, 15] need to observe a large number of similar and dissimilar patches of the same object. Unfortunately, manually extending YT-BB is not feasible for such large number of frames. Thus, we have entertained the possibility of tracker-aided annotation to generate the missing dense bounding box annotations arising between the sparsely occurring original YT-BT ones. State-of-the-art trackers not only achieve impressive performance on standard tracking benchmarks, but they also perform well at high frame rates.

To assess such capability, we conducted four different experiments to decide which tracker would perform best in densely annotating OTB100 [5]. We chose among the following trackers: ECO [12], CSRDCF [44], BACF [30], SiameseFC [14], $\text{STAPLE}_{\text{CA}}$ [31], STAPLE [26], SRDCF [45], SAMF [46], CSK [47], KCF [48], DCF [48] and MOSSE [23]. To mimic the 1-second annotation in TrackingNet Train set, we assume that all videos of OTB100 are captured at 30fps and the OTB100 dataset is split into 1916 smaller sequences of 30 frames. We evaluate the previously highlighted trackers on the 1916 sequences of OTB100 by running them forward and backward through each sequence.

$$\mathbf{x}_{\text{WG}}^{t} = \mathrm{e}^{-\alpha t}\mathbf{x}_{\text{FW}}^{t} + \left(1 - \mathrm{e}^{-\alpha t}\right)\mathbf{x}_{\text{BK}}^{t} \tag{2}$$

The results of both the forward and backward passes are then combined by directly averaging the two results and by generating the convex combination (weighted average) according to Eq. 2, where $\mathbf{x}_{\text{FW}}^{t}$, $\mathbf{x}_{\text{BK}}^{t}$ and $\mathbf{x}_{\text{WG}}^{t}$ are the tracking results at frame $t$ for the forward pass, backward pass, and the weighted average respectively. Note that the maximum sequence length is 30, thus $t \in [1, 30]$. The weighted average gives more weight to the results of the forward pass for frames closer to the first frame and vice verse. $\alpha$ is a constant set to 0.05 for all trackers. Figure 5 shows that most trackers perform almost equally well with the best performance upon using the weighted average strategy. Thereafter, since DCF [48] generates a reasonable accuracy with a frame rate of 300fps, we find it suitable for annotating the large training set in TrackingNet. We run DCF in both a forward and a backward pass where the results of both are later combined in a weighted average fashion as described in Eq. 2.

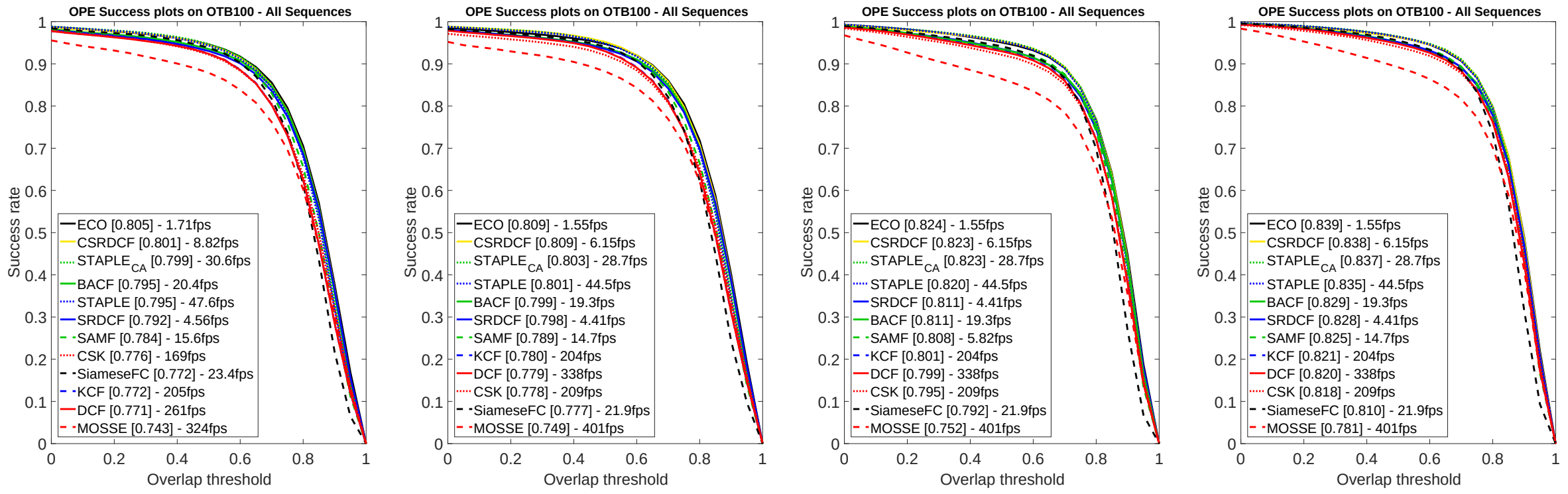

Fig. 5: Tracking results of 12 trackers on the OT100 dataset after splitting it into sequences of length 30 frames. **left to right:** forward pass, backward pass, average results between forward and backward passes, and weighted average as in Eq 2.

# 5 Tracking Benchmark

In our benchmark, we compare a large variety of tracking algorithms that cover all common tracking principles. The majority of current state-of-the-art alogrithms are based on discriminative correlation filters with handcrafted or deep features. We select trackers to cover a large set of combinations of features and kernels. MOSSE [23], CSK [24], DCF [25], KCF [25] use simple features and do not adapt to scale variations. DSST [27], SAMF [28], and STAPLE [26] use more sophisticated features such as Colornames and try to compensate for scale variations. We also include trackers that propose some kind of general framework to improve upon correlation filter tracking. These include SRDCF [49], $\text{SAMF}_{\text{AT}}$ [28], $\text{STAPLE}_{\text{CA}}$ [31], BACF [30] and ECO-HC [12]. We include CFNet [15] and SiameseFC [14] to represent CNN matching trackers and MEEM [21] and DLSSVM [50] for structured SVM based trackers. Last, we include some baseline trackers such as TLD [51], Struck [22], ASLA [19] and IVT [20] for reference. Table 3 summarizes the selected trackers along with their representation scheme, search method, runtime and a generic description.

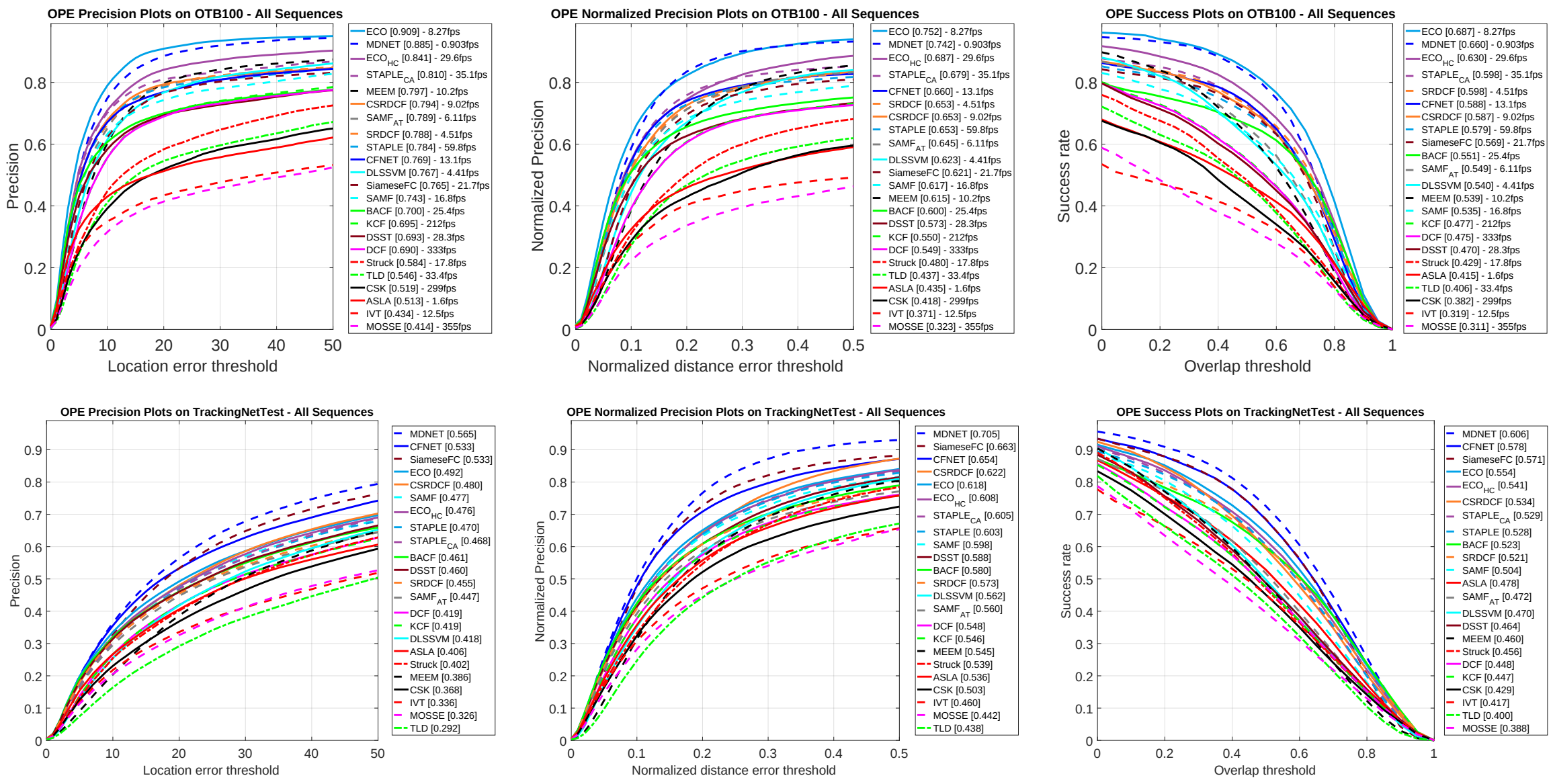

Fig. 6: Benchmark results on OTB100 *top* and on TrackingNet *bottom*.

## 5.1 State-of-the-art Benchmark on TrackingNet

Figure 6 shows the results on the complete dataset. Note that the highest score for any tracker is about 60% success rate compared to around 90% on OTB. The top performing tracker is MD-Net that trains in an online fashion and is, as a result, able to adapt best. However, this comes at the cost of a very slow runtime. Next are CFNet and SiamFC that benefit from being trained on a large-scale dataset (ImageNet Videos). However, as we show later, their performance can be further improved by using our training dataset.

Table 3: Evaluated Trackers. Representation: PI - Pixel Intensity, HOG - Histogram of Oriented Gradients, CN - Color Names, CH - Color Histogram, GK - Gaussian Kernel, K - Keypoints, BP - Binary Pattern, SSVM - Structured Support Vector Machine. Search: PF - Particle Filter, RS - Random Sampling, DS - Dense Sampling.

| **Tracker** | **Representation** | **Search** | **FPS** | **Venue** |
|---|---|---|---|---|
| ASLA[19] | Sparse | PF | 2.13 | CVPR'12 |
| IVT[20] | PCA | PF | 11.7 | IJCVIP'08 |
| Struck[22] | SSVM, Haar | RS | 16.4 | ICCV'11 |
| TLD[51] | BP | RS | 22.9 | PAMI'11 |
| CSK[24] | PI, GK | DS | 127 | ECCV'12 |
| DCF[25] | HOG | DS | 175 | PAMI'15 |
| KCF[25] | HOG, GK | DS | 119 | PAMI'15 |
| MOSSE[23] | PI | DS | 223 | CVPR'10 |
| DSST[27] | PCA-HOG, PI | DS | 11.9 | BMVC'14 |
| SAMF[28] | PI, HOG, CN, GK | DS | 6.61 | ECCVW'14 |
| STAPLE[26] | HOG, CH | DS | 22.1 | CVPR'16 |
| CSRDCF | HOG, CN, PI | DS | 6.17 | IJCV'18 |
| SRDCF[49] | HOG | DS | 3.17 | ICCV'15 |
| BACF | HOG | DS | 12.1 | ICCV'17 |
| ECO_HC[12] | HOG | DS | 21.2 | CVPR'17 |
| SAMF_AT | PI, HOG, CN, GK | DS | 2.1 | ECCV'16 |
| STAPLE_CA[31] | HOG, CH | DS | 15.9 | CVPR'17 |
| CFNET | Deep | DS | 10.7 | CVPR'17 |
| SiameseFC[14] | Deep | DS | 11.6 | ECCVW'16 |
| MDNET[33] | Deep | RS | 0.625 | CVPR'16 |
| ECO[12] | Deep | DS | 4.16 | CVPR'17 |
| MEEM[21] | SSVM | RS | 7.57 | ECCV'14 |
| DLSSVM | SSVM | RS | 5.59 | CVPR'16 |

### 5.2 Real-Time Tracking

For many real applications, tracking is not very useful if it cannot be done at real-time. Therefore, we conduct an experiment to evaluate how well trackers would perform in more realistic settings where frames are skipped if a tracker is too slow. We do this by subsampling the sequence based on each tracker's speed. Figure 7 shows the results of this experiment across the complete dataset. As expected, most trackers that run below real-time degrade. In the worst case, this degradation can be as much as 50%, as is the case for Struck. More recent trackers, in particular deep learning ones, are much less affected. CFNet for example, does not degrade at all even though it only sees every third frame. This is probably due to the fact that it relies on a generic object matching function that was trained on a large-scale dataset.

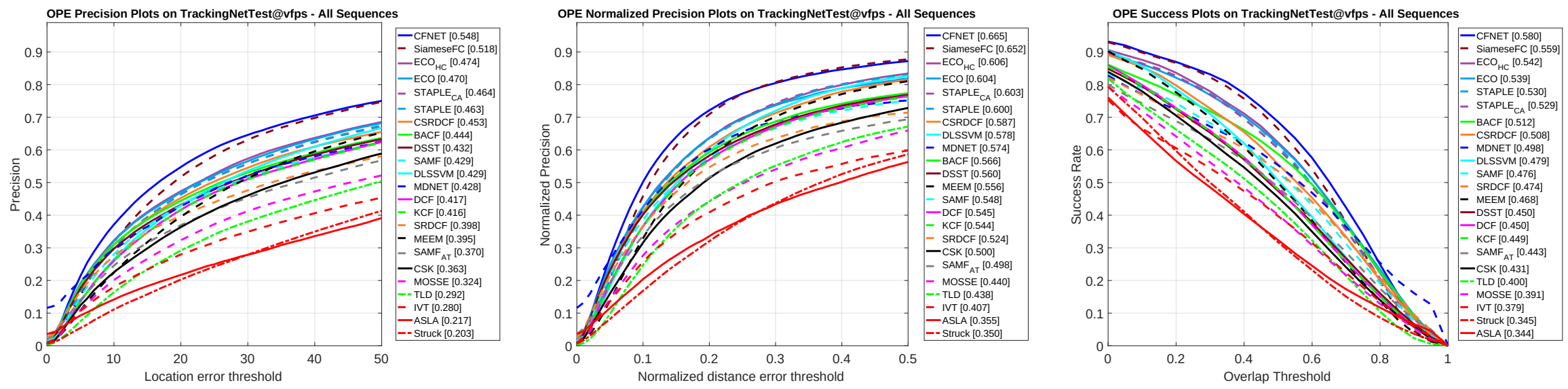

Fig. 7: Benchmark results on TrackingNet with variable frame rate depending on tracker speed.

## 5.3 Retraining on TrainingNet

We fine-tune SiameseFC on a fraction of TrackingNet to show how our data can improve the tracking performance of deep-learning based trackers. The results are shown in Figure 8. By training on only one of the twelve chunks (2511 videos) of our training dataset, we observe an increase in all the metrics on TrackingNet Test and OTB100. The precision increases from 0.533 to 0.543 and from 0.765 to 0.781 respectively. The normalized precision increases from 0.663 to 0.673 and from 0.621 to 0.632 respectively. The success increases from 0.571 to 0.581 and from 0.569 to 0.576 respectively. Fine-tuning using more chunks is expected to improve the performance even further.

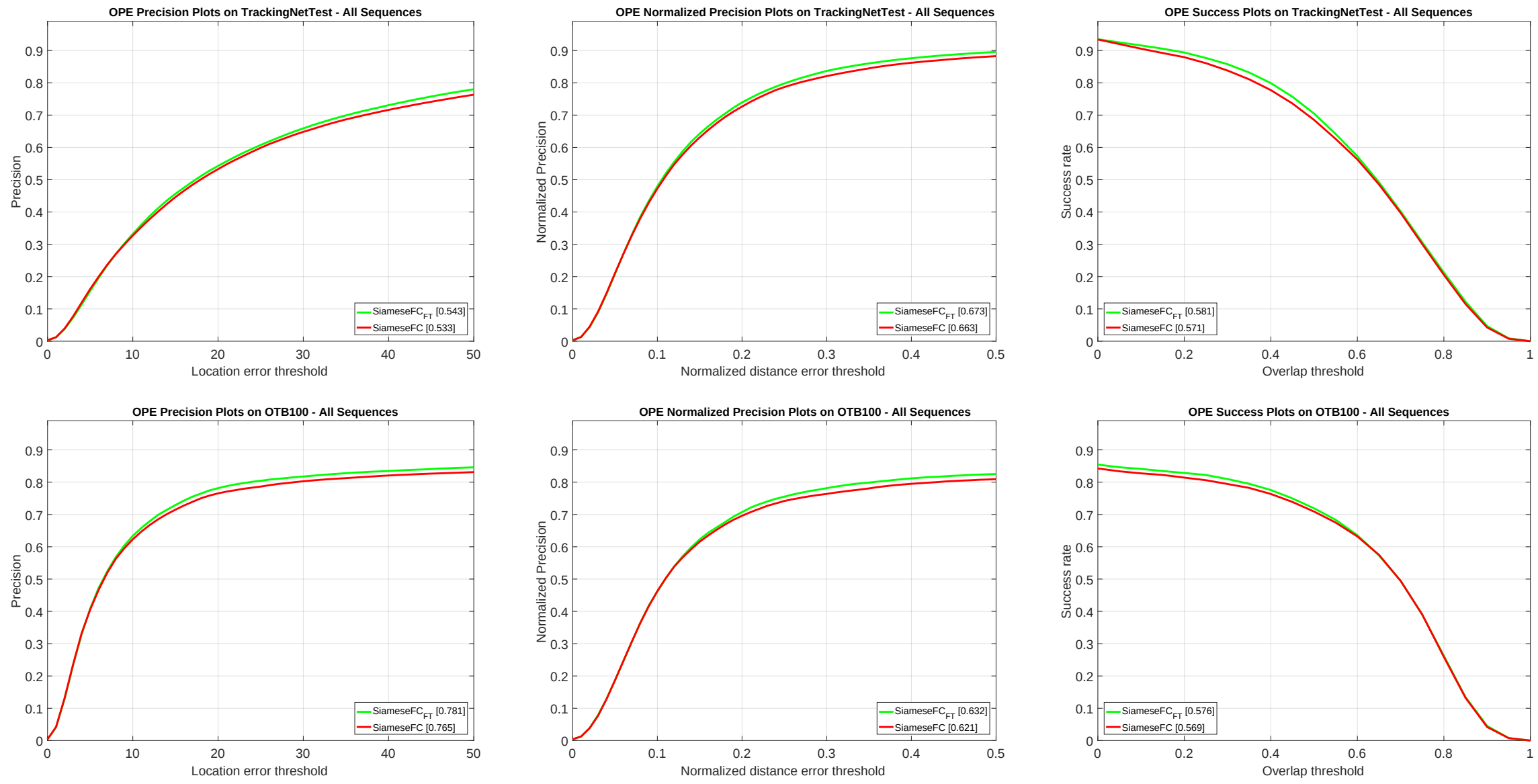

Fig. 8: Fine-tuning results on TrackingNet Test (*top*) and on OTB100 (*bottom*).

### 5.4 Attribute Specific Results

Each video in TrackingNet Test is annotated with 15 attributes described in Section 3. We evaluate all trackers per attribute to get insights about challenges facing state-of-the-art tracking algorithms. We show the most interesting results in Figure 9 and refer the reader to the **supplementary material** for the remaining attributes. We find that videos with in-plane rotation, low resolution targets, and full occlusion are consistently the most difficult. Trackers are least affected by illumination variation, partial occlusion, and object deformation.

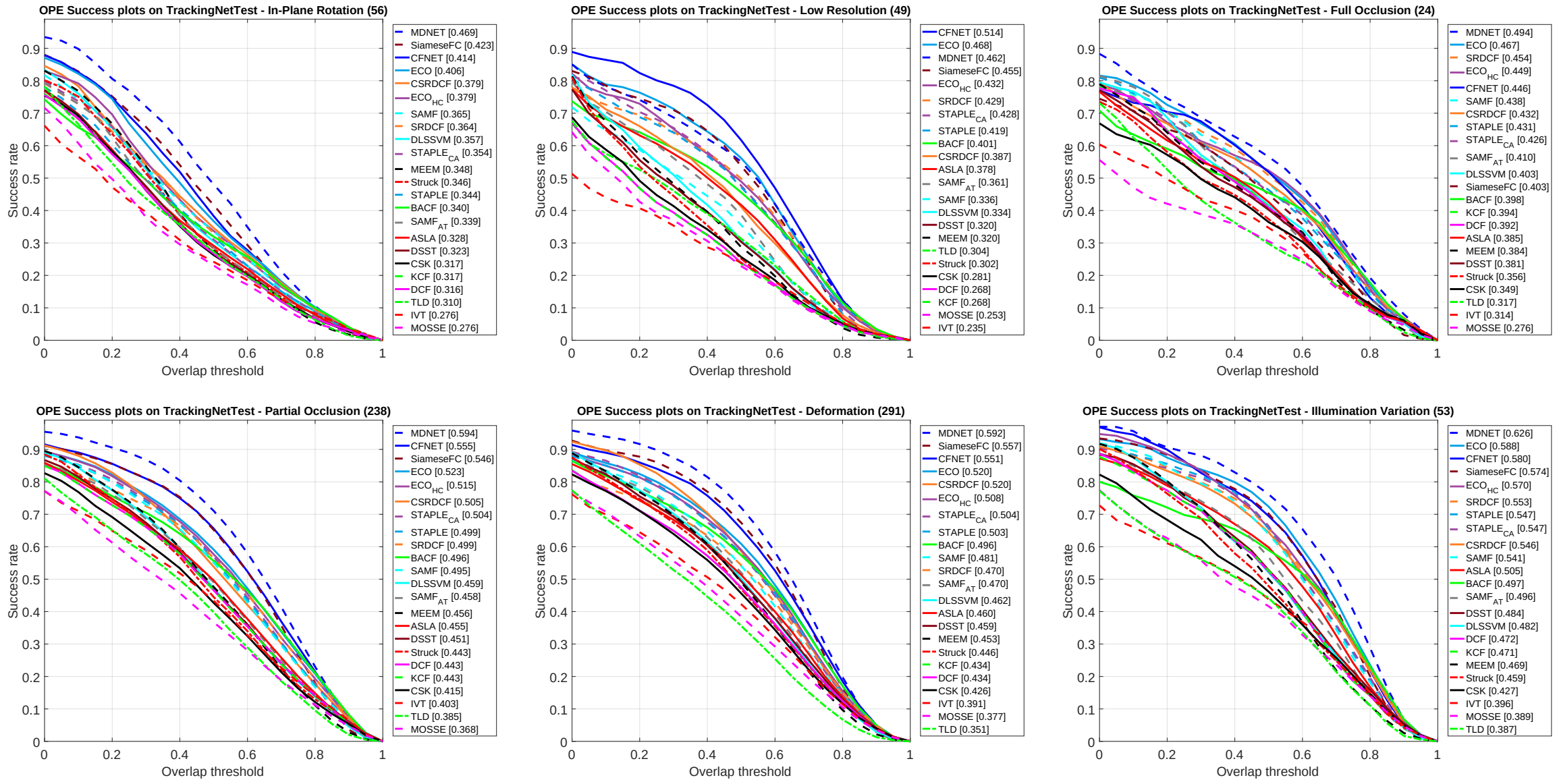

Fig. 9: Per-attribute results on TrackingNet Test.

## 6 Conclusion

In this work, we present TrackingNet, which is, to the best of our knowledge, the largest dataset for object tracking. We show how large-scale existing datasets for object detection can be leveraged for object tracking by a novel interpolation method. We also benchmark more than 20 tracking algorithms on this novel dataset and shed light on what attributes are especially difficult for current trackers. Lastly, we verify the usefulness of our large dataset in improving the performance of some deep learning based trackers.

In the future, we aim to extend the test set from 500 to 1000 videos. We plan to sample the extra 500 videos from different classes within the same category (*e.g.* tortoise / animal). This will allow for further evaluation in regards to generalization. After publication, we plan to release the training set with our interpolated annotations. We will also release the test sequences with initial bounding box annotations and the corresponding integration for the OTB toolkit. At the same time, we will publish our online evaluation server to allow researches to rank their tracking algorithms instantly.